\documentclass{article}

\usepackage{arxiv}

\usepackage{algorithm}
\usepackage{algorithmic}
\usepackage[colorlinks,linkcolor=black,anchorcolor=black,citecolor=black]{hyperref}       
\usepackage{url}            
\usepackage{booktabs}       
\usepackage{amsfonts}       
\usepackage{nicefrac}       
\usepackage{microtype}      
\usepackage{graphicx}
\usepackage{subfigure}
\usepackage{bm}
\usepackage{algorithm,algorithmic}
\usepackage{multirow}
\usepackage{array}
\usepackage{amsmath}

\begin{document}
\title{\Large Dynamic Connected Neural Decision Classifier and Regressor with Dynamic Soft Pruning}
\author{
	Xinyu Fan*
	\\AInnovation Technology Ltd.\\ fanxinyu@ainnovation.com
		}
\date{}
\maketitle

\begin{abstract}
  To deal with various datasets over different complexity, this paper presents an self-adaptive learning model that combines the proposed Dynamic Connected Neural Decision Networks (DNDN) and a new pruning method--Dynamic Soft Pruning (DSP). DNDN is a combination of random forests and deep neural networks that enjoys both the advantages of strong classification capability of tree-like structure and representation learning capability of network structure. Based on Deep Neural Decision Forests (DNDF), this paper adopts an end-to-end training approach by representing the classification distribution with multiple randomly initialized softmax layers, which further allows an ensemble of multiple random forests attached to layers of neural network with different depth. We also propose a soft pruning method DSP to reduce the redundant connections of the network adaptively to avoid over-fitting simple dataset. The model demonstrates no performance loss compared with unpruned models and even higher robustness over different data and feature distribution. Extensive experiments on different datasets demonstrate the superiority of the proposed model over other popular algorithms in solving classification tasks.
\end{abstract}

\section{Introduction}
Machine learning has shown great successes and continues attracting attention in both academic and commercial areas. When dealing with machine learning tasks, related techniques can be divided into two main parts: Deep Neural Networks (DNNs) and Random Forests (RFs). Combining two methods' advantages, this paper presents a generic model that demonstrates high efficiency in learning from datasets of different complexity.

DNNs have been used in many applications, such as image classification\cite{he2016deep} and segmentation\cite{chen2017deeplab}.
However, these methods usually involve a huge number of parameters, which challenge the training and decrease the inference speed. To address these issues, network pruning was proposed to compress the DNN models in order to reduce the model size and time complexity \cite{hassibi1993second}, \cite{lecun1990optimal}, or to delete unimportant weights. For instance, Han et al. \cite{han2015learning} make "lossless" DNN compression by deleting unimportant parameters and retraining the remaining. Guo \cite{NIPS2016_6165} provided dynamic network surgenry with hard pruning. However, those might cause irretrievable network damages thus decreases the learning efficiency and a dynamic soft pruning is needed to imporve the performance.

As one of the most successful ensemble methods, RFs\cite{breiman2001random},     \cite{criminisi2013decision} have shown their efficiency and superiority in many applications. For example, Kontschieder et al \cite{Kontschieder_2015_ICCV} introduced the Deep Neural Decision Forest (DNDF) to facilitate the random forests and DNNs together to further improve the performance. However, the optimization process in DNDF is trained in an iterative way which leads to inefficient and unstable training.

This paper addresses these issues presenting a dynamic neural soft pruning network (DNSPN) that integrates the proposed DNDN with the DSP, where DNDN is a combination of neural networks and random forests and DSP is the newly proposed pruning method. Specifically, in DNDN, we propose a classification distribution with multiple random softmax layers. This enables the optimal predictions for all leaves to be integrated into the optimization of the entire neural network, which allows the whole process to train in an end-to-end way. Also, a more general soft pruning method-DSP is proposed to dynamically prune the weighs of the networks. DSP adopts a soft pruning method to constrain the pruned weights to be in a transition zone where the weights have both the possibilities to be further recovered or cut again. In total, our main contribution can be summarized as follows:

\begin{itemize}
	\item We propose a new model (DNSPN) that integrates DNNs and RFs, requiring low training efforts and demonstrates high learning efficiency in complex datasets.
	\item We present an end-to-end training approach for better combining neural network and random forests (DNDN) and combine different layers to improve the final performance.
	\item A soft pruning method-DSP is proposed to dynamically prune redundant weights while holding high efficiency in recovering the mistakenly pruned weights.
	\item Extensive experiments on several datasets, including UCI Dataset \cite{Dua:2019}, OpenML Dataset\cite{vanschoren2014openml}, AutoML Dataset\cite{guyon2015design}, Self-Designed Dataset, are performed to demonstrate the efficiency of our proposed model compared to other approaches. 
\end{itemize}

The structure of the paper is organized as follows. Section~\ref{related} introduces two related works: DNDF and \textit{Surgery}. Section~\ref{our} presents the details of our proposed DNDN, DSP, Ensemble module together with the DNSPN. Section~\ref{experiment} presents the experiment results of our proposed model on several datasets together with comparisons with other approaches. In the end, Section~\ref{conclusion} concludes our paper and gives several future research directions.

\section{Related Work}\label{related}

\subsection{Deep Neural Networks Combined with Random Forests}

During the past few years much research combined DNNs and RFs. For instance, in \cite{richmond2015relating}, a mapping was explored between the neural networks and RFs and it successfully constructed and initialized a neural network with the trained RF. Bulo et al. \cite{rota2014neural} introduced randomized multi-layer perceptrons to represent each node of the decision tree, which is able to learn the splitting function. The newly proposed Deep Neural Decision Forest (DNDF)\cite{Kontschieder_2015_ICCV} considered the outputs of the top CNN layers as the nodes of the decision tree.

\subsection{Weight Pruning Methods}
Weight pruning aims to cut redundant connections in DNNs and is mainly used in model compression, which can also be used to prevent over-fitting. For practical implementations, Optimal Brain Damage \cite{lecun1990optimal} and Optimal Brain Surgeon \cite{hassibi1993second} perform the weight pruning based on the second-order derivative matrix of the loss function. However, the calculation of the inverse of the Hessian matrix takes additional computation. In the following work, Y. Guo et al. \cite{NIPS2016_6165} proposed \textit{Surgery} which incorporates both the pruning and splicing operations for model compression, where the incorrectly pruned weights are possible to be recovered. In this work, our proposed DSP is inspired by the \textit{Surgery} which will be explained here.

In \textit{Surgery}, taking the k-th layer as an example, the main optimization problem is defined as:
\begin{equation}\label{eq:1}
\min_{\mathbf W_k,\mathbf T_k} L\left(\mathbf W_k\odot \mathbf T_k\right)
\quad \mathrm{s.t.} \ \mathbf T^{(i,j)}_k = \mathbf h_k(\mathbf W^{(i,j)}_k),\forall (i,j) \in \mathcal I
\end{equation}
where $ \mathbf W_k$ represents the weight matrix in the k-th layer and $\mathbf T_k$ is a mask matrix. The symbol $\odot$ refers to the Hadamard product and set $I$ consists of all the entry indices in matrix $W_k$. $\mathbf h_k(\cdot)$ is an indicator function that constructs the $T_k$, which is present as follows:
\begin{equation}\label{eq:4}
\begin{aligned}
\mathbf h_k (\mathbf W^{(i,j)}_k)&=\left\{ \begin{array}{ll}
0&\text{if }0.9\omega >|\mathbf W^{(i,j)}_k| \\
\mathbf T^{(i,j)}_k &\text{if }0.9\omega \leq|\mathbf W^{(i,j)}_k|<1.1\omega  \\
1&\text{if } 1.1\omega \leq |\mathbf W^{(i,j)}_k| \\
\end{array} \right.
\\
&where \quad \omega = (mu+\eta std)
\end{aligned}
\end{equation}
where $\eta$ represents the learning rate, $mu$ and $std$ denote the mean absolute value and the standard deviation of the weights of the k-th layer, respectively.

As shown in Eq. \ref{eq:4}, \textit{Surgery} performed this process in a relatively hard way. In contrast, our proposed DSP performs this in a soft way, which brings in huge advantages for weight recovering.
\begin{figure*}[htp]
	\centering
	\includegraphics[width=0.7\textwidth]{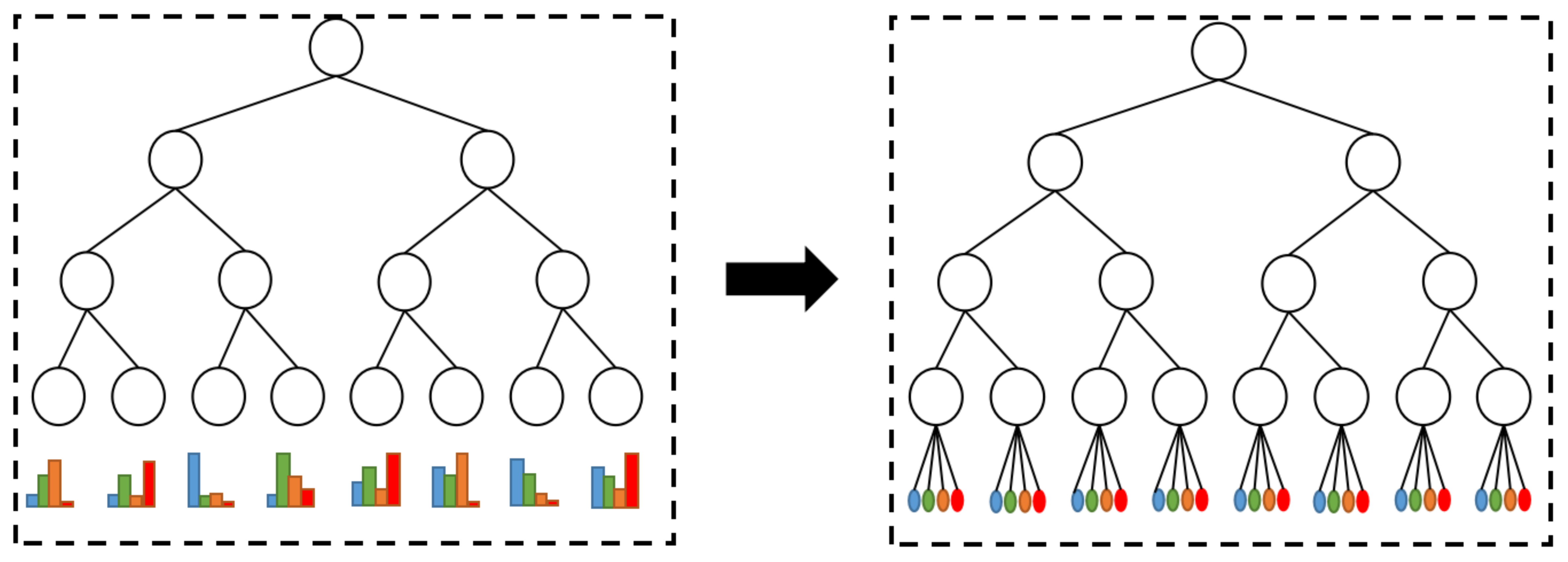}
	\caption{
		The structure comparison between DNDF and our DNDN.
	}\label{DNDN}
\end{figure*}

\section{Proposed Model}\label{our}
Our proposed model DNSPN enjoys the proprieties of representation learning from deep networks and classification ability from decision trees. Together with the dynamic pruning techniques, it is able to learn efficiently from datasets of different complexity. The DNSPN consists of three modules: Dynamic Connected Neural Decision Network (DNDN) module, Dynamic Soft Pruning module, and the Ensemble module. In this section, we will give a brief description of each module and finally show the whole training process of the DNSPN.

\subsection{Dynamic Connected Neural Decision Network}
The DNDF integrates the decision forests with deep neural networks by introducing the stochastic routing into the decision tree. However, learning a decision tree model in DNDF requires estimating both the node parameters $\Theta$ and the leaf predictions $\pi$ in a two-step optimization strategy, which will bring two main problems. First, due to the fixed point iteration optimization method, several forward processes will be performed, which will lead to a slow speed. Second, since the optimization of the leaf node distribution is dependent, the results will become unstable.
To solve this problem, we propose the DNDN which applies the leaf distribution as part of the whole network. This can make the whole process to be trained in an end-to-end way by replacing the original distribution with multiple random initialized softmax layer. An illustration is shown in Figure.\ref{DNDN},

Specifically, we use matrix $\mathbf W_F$ to replace the classification distribution in prediction nodes. The probability of reaching each prediction node from the root is represented by row vector $\mathbf p$. Suppose there are $m$ trees in the forest, each with a depth of h.

For classification, the length of vector $\mathbf p$ is $m\cdot2^{h-1}$ and the size of matrix $\mathbf W_F$ is $(m\cdot2^{h-1} \times k)$ in which k denotes the number of classification labels. Moreover, we use softmax function to ensure that the sum of elements in each row of $\mathbf W_F$ is 1. $\mathbf p$ is constituted by the routing function ${ u }_{ l }(\mathbf x|\bm\Theta )$ mentioned in section 2 and it is given by:
\begin{equation}\label{eq:17}
\mathbf p(\mathbf x|\bm\Theta )=[{ u }_{ { l }_{ 1 } }( \mathbf x|\bm\Theta ),{ u }_{ { l }_{ 2 } }( \mathbf x|\bm\Theta ),..., { u }_{ { l }_{ m\cdot{ 2 }^{ h-1 } } }(\mathbf x|\bm\Theta )].
\end{equation}

The final prediction for classification which is a vector constituted by the probability that the sample belongs to each label is given by:
\begin{equation}\label{eq:18}
output(\mathbf x)=\frac { 1 }{ m } \mathbf p(\mathbf x|\bm\Theta )\cdot { \mathbf W }_{ F }
\end{equation}

Then, we simultaneously update all parameters with a random mini-batch training data by SGD as follows:
\begin{equation}\label{eq:19}
\bm\Theta \leftarrow \bm\Theta -\frac { \eta  }{ | \mathcal{B} | } \sum _{ (\mathbf x,y)\in \mathcal{B}    }^{  }{ \frac { \partial L }{ \partial \bm\Theta  } (\bm\Theta ,\mathbf { W }_{ F };\mathbf x,y) } ,
\end{equation}
\begin{equation}\label{eq:20}
\mathbf { W }_{ F }\leftarrow \mathbf { W }_{ F }-\frac { \eta  }{ | \mathcal{B} | } \sum _{ (\mathbf  x,y)\in  \mathcal{B} }^{  }{ \frac { \partial L }{ \partial { \mathbf W }_{ F } } (\bm\Theta ,\mathbf { W }_{ F }; \mathbf x,y) } ,
\end{equation}

where $L(\bm\Theta ,\mathbf { W }_{ F };\mathbf x,y)$ is the loss function: the cross entropy for classification and the mean square error for regression respectively. $\eta$ is the positive learning rate and $\mathcal{B}\subseteq T$ is a random mini-batch from the training samples. For regression, we can set the size of $\mathbf W_F$ to $(m\cdot2^{h-1} \times 1)$ without additional softmax process. While $\mathbf p$ is obtained as Eq. \ref{eq:17}. The final prediction for regression is a value and calculated by Eq. \ref{eq:18}. The updating of $\mathbf p$ and $\mathbf W_F$ follows formula Eq. \ref{eq:19} and Eq. \ref{eq:20}. It can be seen that the output for regression is the mean of the weighted probability distribution, which endows the regression result with probabilistic significance.

More importantly, the end-to-end training method adopted in DNDN makes it possible to place decision trees after each fully-connected (FC) layer, which significantly increases the model's representation ability.

\subsection{Dynamic Soft Pruning}
This section presents the intuition of our proposed pruning method (DSP) and its implementation details. In particular, we make extensive comparisons between DSP and the traditional approach and explain their differences from a mathematical perspective, demonstrating the high flexibility of the DSP in the pruning process.

\begin{figure}
	\centering
	\includegraphics[width=0.4\textwidth]{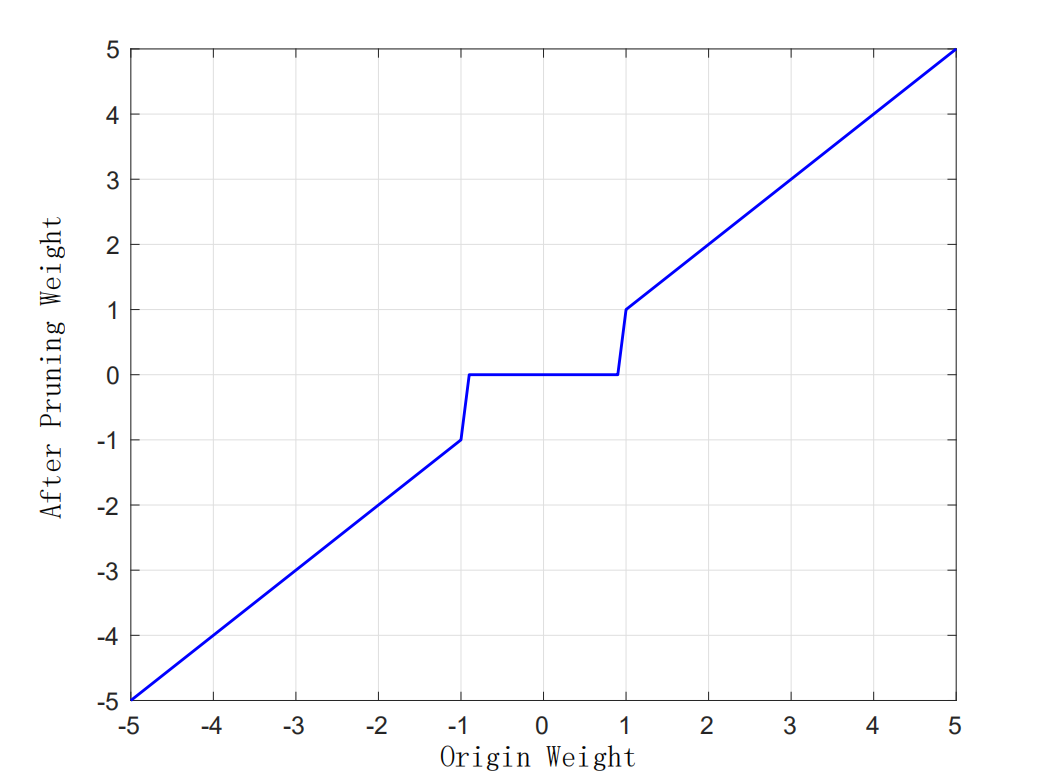}
	\caption{
		The surgery pruning method. The x, y axis represent the original weight and the weight after pruning, respectively.
	}\label{curve1}
\end{figure}

The proposed DSP is inspired by the \textit{Surgery}, an efficient tool for model compression. However, during experiments, several drawbacks in the hard pruning method appear. As shown in Figure.\ref{curve1}, in \textit{Surgery}, there is always a sudden change in the weight value before and after pruning, which will cause information loss and make the training process unstable during training. Besides, weights below the threshold are treated equally, regardless of their distance to the threshold. To tackle these problems, we propose the Dynamic Soft Pruning (DSP) which can prune networks in a soft way. The main idea is to bring in a decaying process to avoid sudden changes in parameters during pruning. We give a detailed description of DSP in the following:

Different from the pruning/splicing function adopted in \textit{Surgery}, as shown in Eq. \ref{eq:4}, DSP proposes a new method to measure the parameter importance:
\begin{equation}\begin{split}\label{eq:16}
\mathbf T^{(i,j)}_k=\mathbf h_k(\mathbf W^{(i,j)}_k)=\max { (\frac { \alpha  }{ \beta  } \cdot { \widetilde {\mathbf T }  }_{ k }^{ (i,j) },{ \widetilde {\mathbf T }  }_{ k }^{ (i,j) }) } ,
\end{split}
\end{equation} 
where ${\widetilde {\mathbf T }  }_{ k }$ is an intermediate function and is presented as:
\begin{equation*}
{ \widetilde {\mathbf T }  }_{ k }^{ (i,j) }=\min { (r ,\beta \log { (\max { (\epsilon ,\frac { |\mathbf W^{(i,j)}_k)| }{ \gamma \cdot mu } ) } ) } ) } .
\end{equation*} 
where, $\alpha$, $\beta$, $\gamma$, $r$, $\epsilon$ are the  hyper-parameters. Specifically, $\alpha$ determines the size of the weight after the soft pruning; $\beta$ is the rate at which the weight decays before it is pruned by soft pruning; $\gamma$ determines the position where the weight will be pruned. The weight is pruned at $\gamma * mu$; $r$ determines the maximum change magnitude of the weight after pruning, where the weight after pruning cannot exceed its original value $r$ times; $\epsilon$ avoids the zero input to the log function.

\begin{figure}
	\centering
	\includegraphics[width=0.4\textwidth]{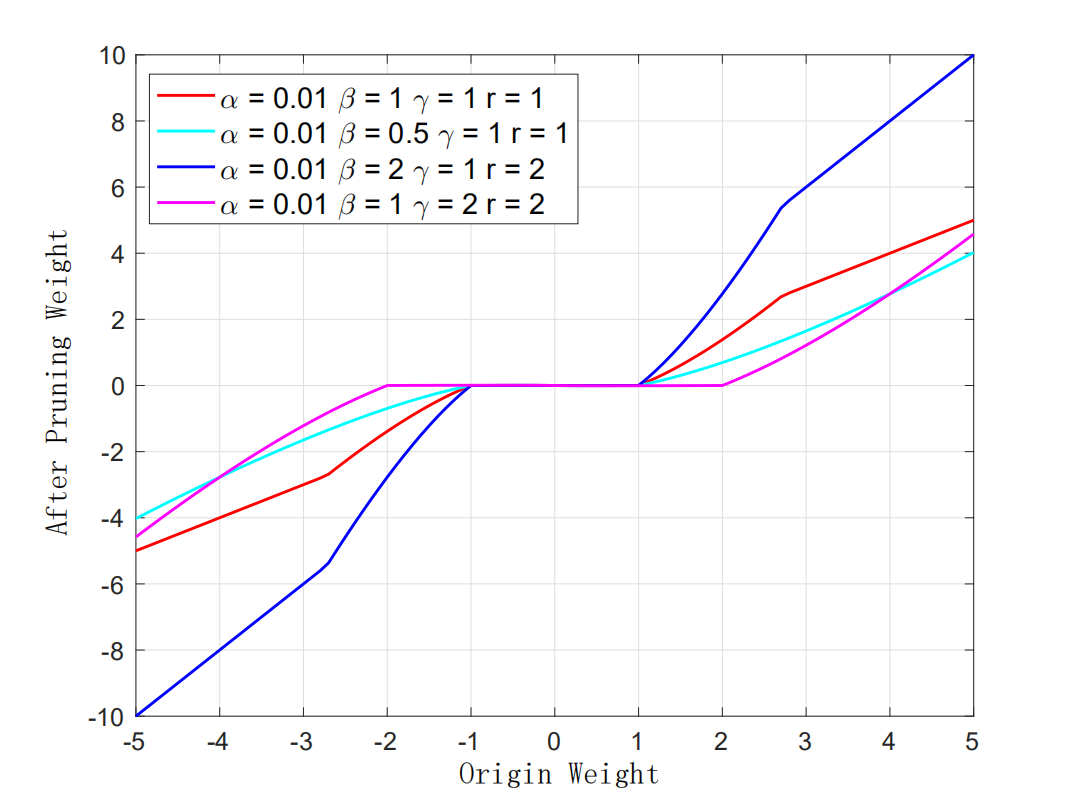}
	\caption{
		The DSP methods under different hyper-parameters configurations. The x, y axis represent the original weight and the weight after pruning, respectively.
	}\label{fig_Tree}
\end{figure}

As shown in Figure.\ref{fig_Tree}, compared to Eq. \ref{eq:4}, which is a multi-stage function, Eq. \ref{eq:16} is a relatively continuous one and the smoothness of its function curve can be adjusted by the hyper-parameters. More importantly, it denotes the underlying difference between \textit{Surgery} and DSP. Our DSP is mainly used for model pruning to prevent over-fitting, where connections are very likely to be pruned or re-established in a high frequency during training, so it is designed to avoid the zero processing, which can increase the recovering efficiency during the future connection recovery. Here, we explain its design principle from a theoretical perspective.
During the recovery process in DSP, let's assume $\gamma$ = 1, given the matrix $\mathbf T_k$, we can update $\mathbf W_k$ by:
\begin{equation}\label{eq:21}
\widetilde {\mathbf { W }}_{ k } = \mathbf  { W }_{ k }\odot \mathbf  { T }_{ k },
\end{equation}
where $\widetilde {\mathbf { W }}_{ k }$ is the actual weight used by the network.

Based on Eq. \ref{eq:21} , we have:
\begin{equation}\label{eq:22}
\frac { \partial \widetilde {\mathbf { W }}_{ k } } { \partial \mathbf  { W }_{ k } } = \frac { \partial { \mathbf { W }_{ k } } * {\mathbf T_k} }{ \partial \mathbf  { W }_{ k } } = \mathbf  T_k + \mathbf W_k * \frac{\partial{\mathbf  T_k}}{\partial{\mathbf W_k}}
\end{equation}
If we set $0 < \alpha < \beta$ in DSP, we have $ \mathbf T_k = \alpha  log(max(\epsilon, \frac{|\mathbf W_k|}{mu}))$.
\begin{equation}\label{eq:22}
\frac { \partial \widetilde {\mathbf { W }}_{ k } } { \partial \mathbf  { W }_{ k } } =
\begin{cases}
\alpha log(\epsilon) &   | \mathbf W_k| \leq \epsilon \cdot mu\\
\alpha \beta (log(\frac{  |\mathbf W_k|}{mu}) + \frac{\mathbf W_k}{|\mathbf W_k|}) & | \mathbf W_k| > \epsilon \cdot mu
\end{cases}
\end{equation}
Where $\alpha$ is usually a small value.Then get the relationship between the gradients before and after pruning:
\begin{equation}\label{eq:25}
\begin{aligned}
\frac{\partial L}{\partial  \mathbf W_k} & = \frac{\partial L}{ \partial \widetilde{\mathbf W_k}} \frac{\partial \widetilde{\mathbf W_k}}{\mathbf W_k} \\
& =\begin{cases}
\alpha log(\epsilon) \cdot \frac{\partial L}{ \partial \widetilde{W_k}} &  |\mathbf W_k| \leq \epsilon \cdot mu\\
\alpha \beta(log(\frac{|\mathbf W_k|}{mu} + \frac{\mathbf W_k}{|\mathbf W_k|})) \cdot \frac{\partial L}{ \partial \widetilde{\mathbf W_k}} & |\mathbf W_k| > \epsilon \cdot mu
\end{cases}
\end{aligned}
\end{equation}
When the pruned weight is relatively small, i.e., $|\mathbf W_k| \leq \epsilon \cdot mu $, we have:
\begin{equation}\label{eq:26}
\frac{\partial L}{\partial \mathbf W_k} = \alpha log(\epsilon) \cdot \frac{\partial L}{ \partial \widetilde{\mathbf W_k}}
\end{equation}
which means that the gradient $\frac{\partial L}{\partial \mathbf W_k}$ will be a relatively moderate value when $\epsilon$ is selected appropriately. Since it is not zero, it is possible to recover $\mathbf W_k$ back to its original value after several iterations.

When $|\mathbf W_k|$ is close to \textit{mu}, we have:
\begin{equation}\label{eq:27}
\begin{aligned}
\frac{\partial L}{\partial \mathbf W_k} &= \alpha\beta (log(\frac{|\mathbf W_k|}{mu} + sign(\mathbf W_k))) \cdot \frac{\partial L}{ \partial \widetilde{\mathbf W_k}} \\
&=  \alpha\beta sign(\mathbf W_k) \cdot \frac{\partial L}{ \partial \widetilde{\mathbf W_k}}
\end{aligned}
\end{equation}
where the gradient is $\frac{\beta }{log(\epsilon)}$ times the value in Eq. \ref{eq:26} and $\beta$ controls the updating rate of $\mathbf W_k$ during the weight recovery.

Based on the soft mechanism, our DSP is able to efficiently recover the pruned weights and put them at a transition zone, where the recovered weights retain both the possibilities of being further recovered or pruned again. While DSP could perform the weight recovering based on Eq. \ref{eq:16}. Besides, in \textit{Surgery}, the recovered weights will be far away from the mu immediately, which means that it is difficult to cut them again. While DSP limits the recovered weights in the vicinity of \textit{mu} so that they preserve the potential to be pruned again, which brings in a more flexible pruning process.

\subsection{Ensemble Module}
Although DNDF has shown excellent performances in classification, it only uses the deeper layers for the tasks without considering the learning representations from other intermediate layers. Considering the fact that low-level features encode precise spatial information and high-level features usually capture the semantic information, many works propose to solve tasks by combining features from different layers. Following this idea, we adopt this strategy to boost the classification performance.

Specifically, based on features from different layers, we get different results $C = {C^1, C^2,..., C^k}$, and our goal is to fuse all these results. Here we exploit the KL divergence-based method to measure the correlation between the original and the ensembled results. The ensembled result $C_{E}$ can be optimized by minimizing the sum of KL divergence as shown in
\begin{equation}
\begin{aligned}
&argmin \sum_{k=1}^KKL(C^k||C_E)  \\
&s.t. \sum q_{i}=1  \\
\end{aligned}
\end{equation}
where $KL(C^k||C_E) = \sum_{i} r_{i}^klog\frac{r_{i}^k}{q_{i}}$ denotes the KL divergence, $ C_E  = (q_1, q_2, ...)$ and  $C^k = (r_1^k, r_2^k, ...)$, the subscript $(i)$ denote the $(i)_{th}$ elements of a vector, where the solution can be deduced by the Lagrange multiplier method.

The final result can be obtained as follows:
\begin{equation}\label{eq:kl2}
C_E = \sum_{k=1}^K\frac{C^K}{K}
\end{equation}
The ensemble module enables our proposed model to deal with different datasets with high flexibility. For example, when dealing with simple datasets, features from the low layers are sufficient to make the final decisions without the necessity to make decisions on the high layers. While dealing with complex datasets, features from the low layers can work as complementary to contribute to making the final decision.

\subsection{Dynamically Neural Soft Pruning Network}
\begin{figure*}
	\centering
	\includegraphics[width=0.8\textwidth]{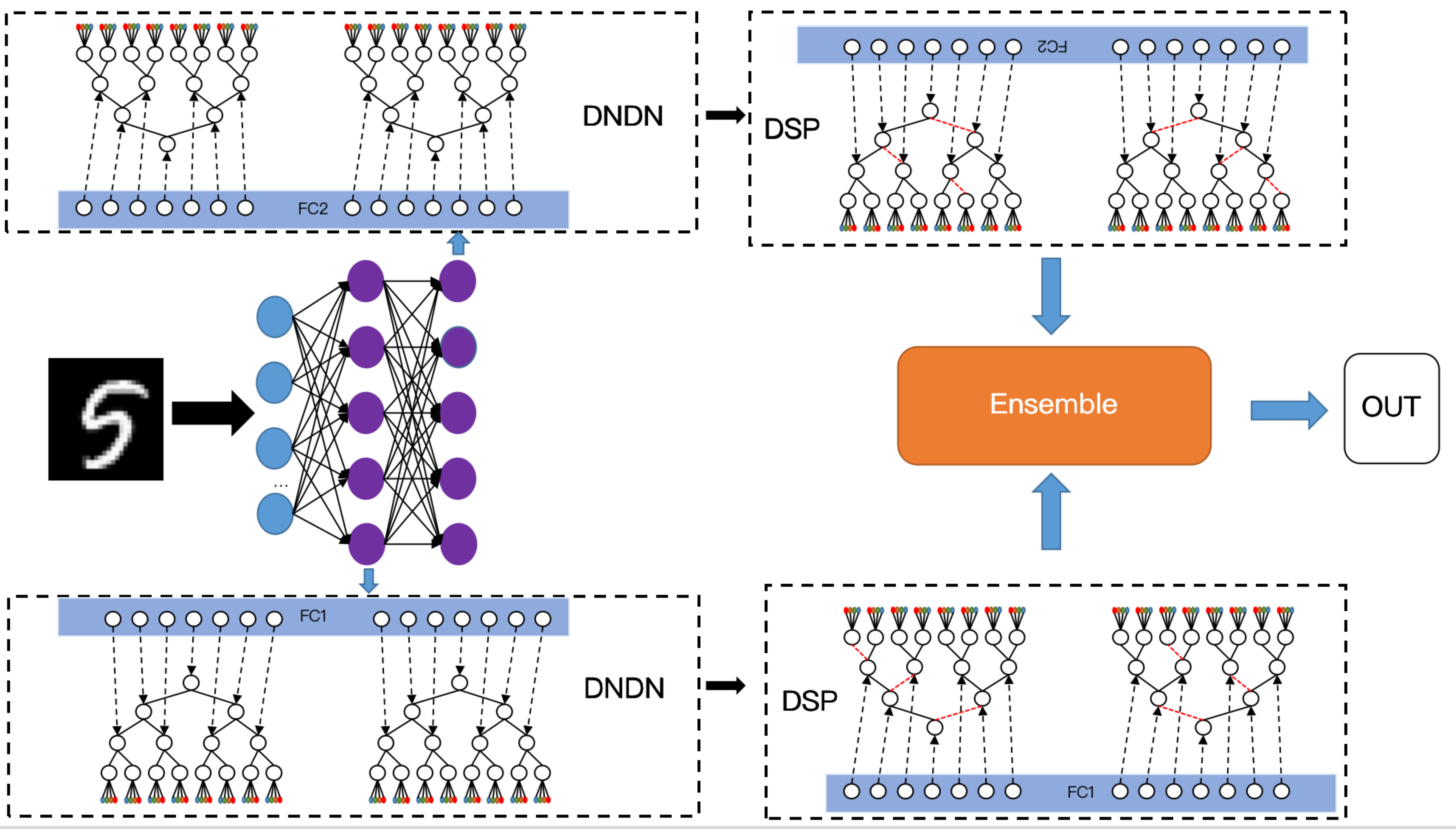}
	\caption{
		The structure of the proposed DNSPN- Dynamic Neural Soft Pruning Network.
	}\label{treeDNN}
\end{figure*}

After illustrating the principles of our proposed model and pruning method, we combine them by embedding DSP into the iterative training process of the DNDN, in an attempt to make full use of their respective advantages. For a classification task, the structure of our proposed model is present in Figure.\ref{treeDNN} and the detailed procedure is shown in Algorithm. \ref{alg:1}.
\begin{algorithm}[h!]
	\caption{Dynamic Connected Neural Decision Classifier and Regressor with soft pruning}
	\label{alg:1}
	\begin{algorithmic}
		\STATE {\bfseries Input:} training dataset D, pruning parameters $\alpha,\beta,\gamma$ and $\delta$\\
		~\\
		\STATE Randomly initialize the parameters $\bm\Theta$ and $\mathbf W_F$ \\
		\REPEAT
		\STATE Choose a minibatch from D
		\STATE Forward propagation and calculate the result with Eq.(\ref{eq:18}) \\
		\STATE Backward propagation and update $\bm\Theta$ with Eq. (\ref{eq:19}) and $\mathbf W_F$ with Eq.(\ref{eq:20}) simultaneously\\
		\STATE Update $\mathbf T_k$ in each layer by current parameters and $\mathbf h(\cdot)$ with Eq.(\ref{eq:16})\\
		\STATE Update $\bm\Theta$ and $\mathbf W_F$ with Eq. (\ref{eq:21}) \\
		\UNTIL{ iterating until satisfying the stop criteria}
	\end{algorithmic}\label{alg:1}
\end{algorithm}
\section{Experiment}\label{experiment}
In this section, we empirically evaluate the performance of the proposed DNSPN. Several state-of-the-art methods are adopted as the baselines, including Decision Tree (DT) \cite{quinlan1986induction}, AutoSkearn\cite{feurer2015efficient}, SVM\cite{scholkopf2001learning}, and the Deep feedforward Fully Connected neural network (DFC) \cite{svozil1997introduction}. Our experiments for exploring the performance and robustness are performed on the four datasets UCI dataset\cite{Dua:2019}, OpenML dataset\cite{vanschoren2014openml}, AutoML dataset\cite{guyon2015design} and our Self-designed Dataset, which we will give a brief introduction below. These four datasets contain more comprehensive machine learning tasks, such as binary classification, 3-classes classification and multi-classes classification from different domains, which can fully explain the superiority of our algorithm.

\subsection{Experimental Setup}
The experiments are performed on the Intel(R) Core(TM) i7-7700 CPU @3.60GHZ, and an NVIDIA Geforce GTX 1080 GPU platform with Tensorflow framework.

For the network, the structure is dynamically designed according to the input data size, we denote the data size with $d$, and the network is designed as Eq.\ref{eq:nn}:
\begin{equation}\label{eq:nn}
d\rightarrow2d\rightarrow2d\rightarrow o
\end{equation}
$o$ represents the output. The learning rate we set here is 1e-3. We use Adam optimization method to train the DNSPN model with batch size equals to 128. The drop-out rate is set to 0.5.

For the forests, the number of the tree is set to 10, and the depth of the tree is 4, which is consistent with \cite{Kontschieder_2015_ICCV}.  The embedding size of the random forest is set with 8.

For the soft pruning, the parameters is set $\alpha=1e-4$, $\beta=1$, $\gamma=1$, $r=1$, $\epsilon=1e-12$.
\subsection{Dataset Introduction}
The experiments are performed on the following datasets to further verify our algorithm. And first we will give a brief introduction to the datasets: OpenML Dataset, UCI Dataset, AutoML Dataset, Self-Designed Dataset.

\textbf{OpenML Dataset}: OpenML \cite{vanschoren2014openml} is a popular database that has been used by many machine learning researchers. The data in OpenML includes much detailed information and can be organized effectively to deal with different learning tasks.

\textbf{UCI Dataset}: The UCI Machine Learning Repository \cite{Dua:2019} is a collection of databases, domain theories, and data generators that are used by the machine learning community for the empirical analysis of machine learning algorithms.

\textbf{AutoML Dataset}: As introduced in \cite{guyon2015design}, the dataset used in AutoML challenge is a dataset formatted in fixed-length feature-vector representations which can be used to verify the regression or classification algorithm (binary, multi-class, or multi-label). The datasets are from a wide range of applications, such as biology, ecology, and so on. All datasets are presented in the form of data matrices with samples in rows and features in columns.

\textbf{Self-Designed Dataset}:  The Self-Designed Dataset is designed to test the robustness of the algorithm. To test the ability of the model to select features and resist noise in simple datasets with high noise, we construct Linear-k dataset.  In addition, the Quadratic-k datasets are constructed to test the ability of the model to pick features and resist noise in complex datasets with high noise (involving the square of features).

For all the datasets, the $N$ represents the dimensions of features, the $Ptr$ represents the number of training samples.

\subsection{Experimental Result and Analysis}
Several experimental results are shown to compare our DNSPN with other methods on the four datasets.
\subsubsection{OpenML Dataset}
The AUC is computed to further demonstrate the superiority of our results compared with other famous algorithms, such as Auto-Sklearn \cite{feurer2015efficient}, FCNN (Fully Connected Neural Networks)\cite{svozil1997introduction} and LightGBM\cite{ke2017lightgbm}.

\begin{figure}[h!]
	\centering
	\includegraphics[width=0.4\textwidth]{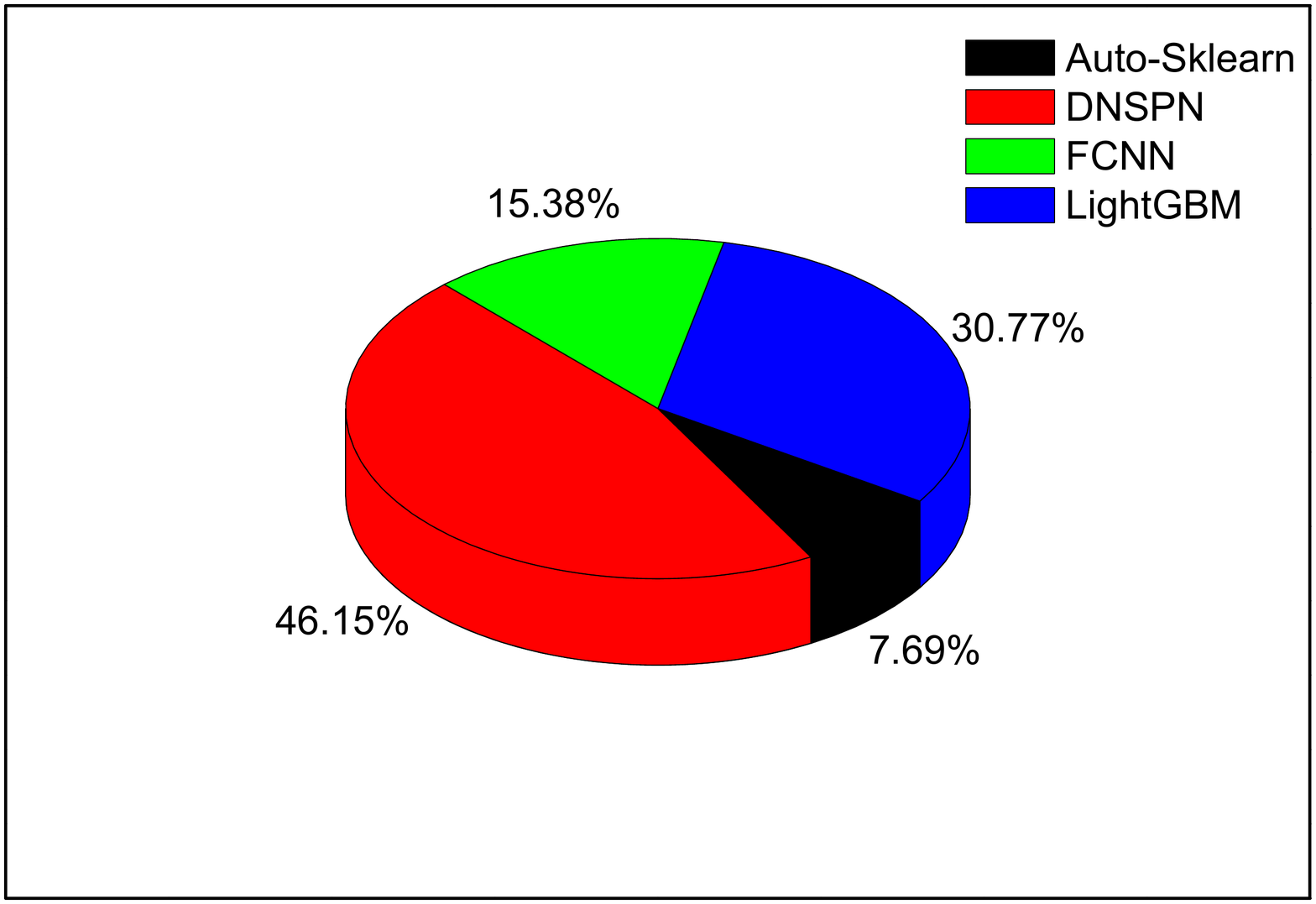}
	\caption{The proportion of optimal results achieved by different models}\label{figOL}
\end{figure}

The AUC is the area under the ROC curve, and the results can be seen in Tab. \ref{tab:3}. To better illustrate the advantages of our algorithm, as can be seen in Fig.\ref{figOL}, our DNSPN has performed well on most of the datasets.

As shown in Tab. \ref{tab:3}, we achieved the best results in 6 out of the 13 datasets. Compared with FCNN which can be seen as our baseline, our DNSPN has a great improvement on the prediction accuracy among 12/13 of the datasets. Due to two reasons: First, since our DNSPN has adopted the DSP module, the model can be fully trained without overfitting and   learn the most relevant features in the training datasets. Second, the ensemble module takes full advantage of the semantic and detailed information contained in the network. Finally, as DNSPN can be trained in an end-to-end way, the training time can be reduced by 40 \%. While for Auto-Sklearn and LightGBM, a little more accuracy was improved at the cost of longer training time.

\begin{table*}[h!]
	\centering
	\caption{The comparison of our DNSPN with other four famous datasets conduced on OpenML DataSet measured with AUC.}
	\begin{tabular}{cccccccc}
		
		\toprule
		Dataset ID&Task& N& Ptr &Auto-Sklearn & FCNN& LightGBM &DNSPN\\
		\midrule  
		38&Binary&29&3395&99.41	&98.83$\pm$0.47&\textbf{99.85$\pm$0.082} &99.11$\pm$0.39\\
		46&3-classes&60&2871&99.46	&99.34$\pm$0.17&99.49$\pm$0.26 &\textbf{99.56$\pm$0.28}\\
		179&Binary&14&43958&87.14 &91.34$\pm$0.33&91.36$\pm$0.29 &\textbf{91.46$\pm$0.34}\\
		184&18-classes&6&25250&98.51 &99.21 $\pm$0.078&97.42 $\pm$1.65 &\textbf{99.28 $\pm$0.12}\\
		389&17-classes&2000&2217&\textbf{99.04} &	98.09$\pm$0.57&98.90$\pm$0.23 &98.26$\pm$0.4869\\
		554&10-classes&784&63000&99.88	&99.96$\pm$0.01&99.97$\pm$0.005 &\textbf{99.97$\pm$0.005}\\
		772&Binary&3&1960&50.92 &\textbf{52.75$\pm$3.62}&51.27$\pm$2.19 &49.91$\pm$3.85\\
		917&Binary&25&900&90.7 &71.05$\pm$4.17&\textbf{97.43$\pm$1.09} &73.78$\pm$4.13\\
		1049&Binary&37&1312&93.13 &91.23$\pm$2.93	&\textbf{93.35$\pm$2.55} &91.23$\pm$3.36\\
		1111&Binary&230&45000&72.36 &78.35 $\pm$1.9436&\textbf{80.31$\pm$1.61} &78.36$\pm$1.79\\
		1120&Binary&10&17118&93.02 &93.29$\pm$ 0.42&93.72$\pm$0.42 &\textbf{93.80$\pm$0.51}\\
		1128&Binary&10935&1390&97.16&\textbf{98.15 $\pm$1.40}&97.88$\pm$1.65 &97.49$\pm$1.85\\
		293&Binary&54&522911&99.35&\textbf{99.55$\pm$0.077}&99.48$\pm$0.047&98.41$\pm$0.088\\
		
		\bottomrule
	\end{tabular}
	\label{tab:3}
\end{table*}

\subsubsection{UCI Dataset}
The experimental results conducted on 9 UCI datasets are shown in Tab \ref{tab:1} compared with Decision Tree (DT), SVM, Auto-Sklearn, and FCNN.

\begin{table*}[htbp]
	\centering
	
	\caption{The comparison of our DNSPN with DT, SVM, Auto-Sklearn, FCNN conduced on UCI datasets}
	\begin{tabular}{cccccccc}
		
		\toprule
		Data Names& N& Ptr &DT & SVM &Auto-Sklearn &FCNN & DNSPN\\
		\midrule  
		Cardiotocography &10 &2.10k&89.93&88.57&91.78& \textbf{97.88$\pm$0.068} &97.71$\pm$0.045\\
		Breast Cancer &30 &0.57k&91.72&96.82&97.28&97.89$\pm$0.008 &\textbf{99.38$\pm$0.006}\\
		Nomao&120&34.5k&93.47&94.35&96.39&97.13$\pm$0.003& \textbf{99.10$\pm$0.001}\\
		
		Multiple Features&216&2.00k&19.50&20.60&22.00&97.04$\pm$0.053&\textbf{97.88$\pm$0.019}\\
		Arrhythmia&279&0.45k&48.68&56.81&56.04&62.24$\pm$0.033&\textbf{64.25$\pm$0.057}\\
		Madelon&500&2.60k&73.67&58.445&\textbf{82.78}&60.47$\pm$0.042&66.03$\pm$0.022\\
		SECOM&591&1.57k&54.71&57.74&50.00&67.72$\pm$0.040&\textbf{69.31$\pm$0.045}\\
		ISOLET5&617&1.56k&\textsl{5.83}&\textsl{7.821}&\textsl{7.37}&94.21$\pm$0.023&\textbf{95.93$\pm$0.042}\\
		Gisette&5000&7.00k&93.08&97.40&97.64&98.89$\pm$0.003&\textbf{99.58$\pm$0.002}\\
		\bottomrule
	\end{tabular}
	\label{tab:1}
\end{table*}

The results show that our algorithm performs well on most of the datasets. In ISOLET5 dataset, the traditional machine learning algorithms such as DT, SVM, Auto-Sklearn performs poorly mainly due to a large number of categories and weak features. For the Madelon dataset, both the results of our DNSPN and FCNN are not ideal because the original dataset has a large number of redundant features. However, the number of the training sample is not big enough to make the pruned module to learn good features, which will make the model fall into over-fitting.

\subsubsection{AutoML Dataset}
Another famous dataset chosen here is AutoML DataSet. And the experimental results is shown in Tab. \ref{tab:4}.
\begin{table*}[h!]
	\centering
	\caption{The comparison of our DNSPPN with  Auto-Sklearn and FCNN  conduced on AutoML DataSet.}
	\begin{tabular}{ccccccc}
		\toprule
		name&Task& N& Ptr &Auto-Sklearn &FCNN& DNSPN\\
		\midrule  
		Albert&Binary&78&382716&66.34&65.84$\pm$0.034&\textbf{77.31$\pm$0.006}\\
		Credit&5-classes&2000&9000&97.20&96.04$\pm$0.059&\textbf{97.72$\pm$0.035}\\
		Dilbert&7-classes&800&7413&64.20&67.21$\pm$0.121&\textbf{67.79$\pm$0.035}\\
		Helena&4-classes&54&75360&69.60&71.67$\pm$0.029&\textbf{72.17$\pm$0.037}\\
		Jannis&10-classes&180&52479&\textbf{69.01}&67.01$\pm$0.070&68.92$\pm$0.094\\	
		\bottomrule
	\end{tabular}
	\label{tab:4}
\end{table*}

As can be seen in Tab. \ref{tab:4}, our DNSPN performed well on 4 of the 5 datasets. For the Jannis dataset, due to the poor performance of the FCNN, although the DNSPN has a great improvement of FCNN, the DNSPN does not perform as well as Auto-Sklearn.

\subsubsection{Self-Designed Dataset with noise}
First, a detail description about the dataset will be given. The original data is a 100-dimensional, dimensionally-independent data sampled from a standard normal distribution, which can be seen in Eq. \ref{eq:dd1}
\begin{equation}\label{eq:dd1}
x  = (x_1, x_2, ..., x_{100})^T,~~~x_i\sim N(0,1)
\end{equation}

And the noise is added to the original data to get the noise dataset as can be seen in Eq. \ref{eq:dd2}
\begin{equation}\label{eq:dd2}
x_i\rightarrow x_i+\epsilon_i,~~~~\epsilon_i\sim N(0,\sigma_i^2)
\end{equation}
where $\sigma_i$ demonstrates the amount of noise, $\sigma_i=0$ means no noise.

\begin{table*}[h!]
	\centering
	
	\caption{The result for noise resistance between FCNN, DNDF and DNSPN}
	\begin{tabular}{ccccccc}
		
		\toprule
		dataset&FCNN&DNDF& DNSPN\\
		\midrule  
		Linear-5&93.57$\pm$0.041&93.78$\pm$0.048&\textbf{96.10$\pm$0.018}\\
		Linear-50&97.59$\pm$0.003&97.62$\pm$0.003&\textbf{98.45$\pm$0.002}\\
		Quadratic-5&92.61$\pm$0.025&93.01$\pm$0.031&\textbf{96.35$\pm$0.014}\\
		Quadratic-50&87.57$\pm$0.023&88.26$\pm$0.034&\textbf{92.96$\pm$0.018}\\
		Quadratic-50(noise less)&93.89$\pm$0.010&94.31$\pm$0.004&\textbf{94.90$\pm$0.008}\\
		\bottomrule
	\end{tabular}
	\label{tab:5}
\end{table*}

Then we choose $k$ dimensions randomly from these 100 dimensions as really useful features.
\begin{equation}\label{eq:dd3}
x\rightarrow\tilde{x}=(x_{i_1},x_{i_2},...,x_{i_k})^T
\end{equation}

The linear-k means the linear transformations. We choose
$k+1$ samples from the standard normal distribution to be the weight and bias of the linear map, and the label is determined by whether the linear map is greater than 0 as can be seen in Eq. \ref{eq:dd4}.
\begin{equation}\label{eq:dd4}
y=\text{sign}(w^T\tilde x+b)
\end{equation}
For the Quadratic-k dataset, choosing k dimensions randomly from these 100 dimensions twice as the features.
\begin{equation}
\begin{aligned}
&x\rightarrow\tilde{x}_1=(x_{i_1},x_{i_2},...,x_{i_k})^T \\
&x\rightarrow\tilde{x}_2=(x_{j_1},x_{j_2},...,x_{j_k})^T
\end{aligned}\label{eq:dd5}
\end{equation}
Then 2k+1 samples were collected from the standard normal distribution to be used as the $w_1$, $w_2$ and $b$ of the quadratic mapping, and the label was determined by whether the quadratic mapping was greater than 0, as can be seen in Eq. \ref{eq:dd5}
\begin{equation}\label{eq:dd5}
y=\text{sign}(w_1^T\tilde x_1^2+w_2^T\tilde x_2+b)
\end{equation}
The experiment results of FCNN, DNDF and DNSPN are shown Tab. \ref{tab:5}.

As shown in Tab. \ref{tab:5}, our DNSPN performs better than FCNN and DNDF in all the datasets, demonstrating the efficiency of DNSPN to select features and resist the noise. With the DSP module, unimportant weights in the network will be pruned, which can greatly reduce the redundancy of the network and increase the robustness despite the dimensions of input features, different mapping of the input features, the amount of the noise added to the input features. Firstly, for the Linear-5 and Linear-50 dataset with the same noise, for the higher dimensional dataset, FCNN, DNDF, and DNSPN perform well. When the dimension drops, the little noise could lead to large changes in data distribution, the accuracy of FCNN and DNDF drops a lot, our DNSPN still keeps the higher accuracy. This means that our DNSPN can resist the noise while ignoring the dimensions of the features.
Secondly, for the Linear-50 and Quadratic-50 dataset, with the mapping from linear to quadratic, the accuracy of FCNN and DNDF drops a lot, to almost ten percent, while DNSPN still keeps the high accuracy. In other words, our DNSPN can easily handle the problem whether it is linear mapping or quadratic mapping, while the FCNN and DNDF can only deal with the linear mapping problem. Thirdly, for the Quadratic-50 and Qudratic-50 with less noise, with the increase of the noise, the performances of FCNN and DNDF drop quickly, while the DNSPN still keeps the high accuracy. In other words, the other two algorithms can only solve the problem with little noise, while our DNSPN can resist more noise.

\subsection{Result Analysis}
In the experiments above, the DNSPN has shown excellent performance, which mainly benefits from two aspects. First, the training method of the original DNDF is modified into end-to-end training, which greatly reduces the training efforts. Also, the traditional DNDF structure is improved by adding DNDF units after each FC layer, allowing the model to deal with complex datasets.
Second, with the use of pruning, the network redundancy is reduced greatly and the over-fitting problem is mitigated to some extent. Especially, compared to the hard pruning, the proposed DSP adopts the soft pruning enables a gradually changing phase. This brings in a flexible pruning phase and is also beneficial to the training of the neural network. Furthermore, soft pruning makes the mappings from feature layers to nodes in DNDF more characteristic, which contributes to the final predictions.

\section{Conclusion}\label{conclusion} 
In this paper, we propose DNSPN which integrates DNDN, DSP and Embedding module together to efficiently learn from datasets of different complexities. For the DNDN module, it improves the structure used in DNDF\cite{Kontschieder_2015_ICCV} and proposes an end-to-end training approach, which can improve the stability and speed of the training. In addition, the DSP is proposed to cut redundant connections in networks to avoid the over-fitting problem and improve robustness. Furthermore, different layers are embedded together to combine the detailed and semantic learning representations for the final prediction, which can further improve the accuracy and robustness of our DNSPN. Finally, experimental results performed on  UCI, OpenML, AutoML, Self-Constructed datasets demonstrate the superiority of DNSPN both from the perspective of robustness and accuracy.

Despite the learning efficiency in dealing with complex datasets, our DNSPN shows the limited capability of learning from datasets with few samples, due to the limitation of training methods of deep neural networks. Our future work will tackle the few-sample learning problem.

\bibliographystyle{my}
\bibliography{references}
\end{document}